\documentclass[a4paper,10pt,dvipdfm]{article}

\usepackage{laarstyle}
\usepackage{amsmath,amssymb}
\usepackage{latexsym}
\usepackage{graphicx}
\usepackage[tight]{subfigure}
\usepackage{rotating}
\usepackage{float}
\usepackage{url}
\usepackage[abbr]{harvard}
\usepackage[active]{srcltx}
\usepackage{flushend}
\usepackage{pb-diagram}
\begin{document} 

\title{ON THE USE OF LEE'S PROTOCOL FOR SPECKLE-REDUCING TECHNIQUES}

\author{E.\ MOSCHETTI$^{(1)}$, M.\ G.\ PALACIO$^{(1)}$, M.\ PICCO$^{(1)}$, O.\ H.\ BUSTOS$^{(2)}$ 
and A.\ C.\ FRERY$^{(3)}$}

\affiliation{
$^{(1)}$Departamento de Matem\'atica\\
Facultad de Ciencias Exactas F\'\i sico Qu\'\i mica y Naturales\\
Universidad Nacional de R\'\i o Cuarto\\
Ruta 36 km 601, X5804BYA R\'\i o Cuarto -- Argentina \\
\texttt{\{emoschetti;gpalacio;mpicco\}@exa.unrc.edu.ar}\\
$^{(2)}$Facultad de Matem\'atica, Astronom\'\i a y F\'\i sica\\
Universidad Nacional de C\'ordoba\\
Ing.\ Medina Allende esq.\ Haya de la Torre, 5000 C\'ordoba -- Argentina\\
\texttt{obustos@arnet.com.ar}\\
$^{(3)}$Instituto de Computa\c c\~ao\\
Universidade Federal de Alagoas\\
57072-970 Macei\'o, AL -- Brasil\\
\texttt{acfrery@pesquisador.cnpq.br}}

\maketitle



\abstract
This paper presents two new MAP (Maximum a Posteriori) filters for speckle noise reduction and a Monte Carlo procedure for the assessment of their performance.
In order to quantitatively evaluate the results obtained using these new filters, with respect to classical ones, a Monte Carlo extension of Lee's protocol is proposed.
This extension of the protocol shows that its original version leads to inconsistencies that hamper its use as a general procedure for filter assessment.
Some solutions for these inconsistencies are proposed, and a consistent comparison of speckle-reducing filters is provided.
\endabstract

 
\section{INTRODUCTION}

Contemporary remote sensing relies on data from different regions of the electromagnetic spectrum, the optical, infrared, and microwaves ones.
Synthetic Aperture Radar (SAR) sensors are becoming more relevant in every field of research and development that employs remotely sensed data, since they are active and, thus, not requiring external sources of illumination.
They can observe the environment in a wavelenght that is little or not affected at all by weather conditions, providing complementary information to the conventional optical sensors.
The information these sensors provide is relevant for every remote sensing application, including environmental studies, anthropic activities, oil spill monitoring, disaster assessment, reconnaissance, surveillance and targeting, among others.

As every image obtained using coherent illumination, as is the case of laser, sonar and ultrasound-B imaging, SAR images suffers from speckle noise.
Such type of noise do not follows classical Gaussian additive, it is multiplicative noise.
Classical techniques for noise reduction are thus inefficient to combat speckle see, for instance, ~\citeasnoun{AllendePRL:01}; ~\citeasnoun {DelignonPieczynski:02}; ~\citeasnoun{Kuttikkad00};~\citeasnoun {MedeirosMascaCosta:03} ;~\citeasnoun{Touzi:02:reviewSARfiltering}   .

Since speckle noise hampers the ability to identify objects, many techniques have been proposed to alleviate this issue.
Techniques are applied during the generation phase of the images \citeaffixed{lolane}{multilook processing, see} or after the data is available to the users (processing with filters).
A ``good'' technique must combat speckle and, at the same time, preserve details as well as relevant information.

In order to assess the performance of speckle-reduction techniques (multilook or filter-based),  \citeasnoun{leejurkevich94}  proposed a protocol. 
It consists of a phantom image corrupted by speckle noise processed by speckle-reduction tecniques.
Measures of quality are computed on the images obtained, and the performance of the used tecnhique is assessed from these measures.
This protocol can be applied to both multilook or filter-based speckle-reduction procedures.
In this paper we will discuss the use of this protocol, termed ``Lee's protocol'', on filter-based techniques.

This papers presents situations where Lee's protocol is inadequate and should be replaced by Monte Carlo experiments; the outline of such simulation is presented.
This approach aims at results that are representative for a collection of images, while the ones provided by Lee's protocol regard only one image and, as will be shown, can be biased and uninformative.

Among the many approaches for speckle-reduction using filters, one should mention those based on the noise statistical properties and other general purpose techniques (median, mean etc.).
For a comprehensive review of speckle filters the reader is referred to the works by  \citeasnoun{leejurkevich94}, by \citeasnoun{MedeirosMascaCosta:03} and to the references therein.

Among others, \citeasnoun{geman84} show the advantage of using stochastic models in image processing.
They propose a general transformation setup:
$$
\begin{diagram}
\node{X} \arrow{e,t}{\tau} \node{Z} \arrow{e} \node{\widehat X}
\end{diagram}
$$
where $X$ represents the unobserved true data, $\tau$ transformations imposed by the sensor and $Z$ the observed data.
The sensor typically degrades the truth by means of non-linear and non-invertible transformations and with the inclusion of noise.
The knowledge of the properties of both $X$ and $\tau$ allows building techniques for obtaining $\widehat X$, an estimate of $X$.

With the above setup and a Bayesian approach, many estimators can be used to compute $\widehat X$ as, for instance, maximum likelihood, minimum squared error, maximum posterior mode and maximum a posteriori (MAP).
One of the contributions of this paper is the proposal of two new MAP estimators for the ground truth of SAR imagery, and the assessment of their performance.

Image quality assessment in general, and filter performance evaluation in particular, are hard tasks~\cite{WangBovikLu:ICASSP:02}.
Many factors are involved as, for instance, the true scene, the degradation and the type of application sought for the data.
Lee's protocol is a proposal for filter assessment based on the extraction of measures of quality; in its original version the performance of a filter is assessed using a single image $\widehat X$ as input.
The use of statistical models allows the proposal of Monte Carlo experiments in order to compute quantities that, as noise reduction performance, depend on many factors and are hard (or impossible) to obtain directly (see, for instance, Bustos and Frery,1992; Robert and Casella, 2000).
We show here that Lee's protocol requires such a Monte Carlo procedure in order to precisely assess the performance of speckle filters.

The paper is organized as follows.
Section~II presents the multiplicative model.
In Section~III two new speckle filters based on a Bayesian approach on this model are proposed, after reviewing one of the most widely used techniques: Lee's filter~\citeaffixed{Lee86}{proposed by}.
Section~IV presents Lee's protocol, and Section~V describes a Monte Carlo experiment used to assess the new filters with respect to the one proposed by~\citeasnoun{leejurkevich94}; in this section we see that the original protocol is inadequate for filter assessment.
Sections~VI and~VII present the results and conclusions.

\section{THE MULTIPLICATIVE MODEL}\label{sec:multi}

Only univariate signals will be discussed here; the reader interested in multivariate SAR statistical modelling is referred to~\citeasnoun{FreitasFreryCorreia:Environmetrics:03}.

\citeasnoun{goodman85} provided one of the first rigorous statistical frameworks, known as ``Multiplicative Model'' for dealing with speckle noise in the context of laser imaging.
The use of such framework has led to the most successful techniques for SAR data processing and analysis.
This phenomenological model states that the observation in every pixel is the outcome of a random variable $Z\colon\Omega\rightarrow\mathbb R_+$ that, in turn, is the product of two independent random variables: $X\colon\Omega\rightarrow\mathbb R_+$, the ground truth or backscatter, related to the intrinsic dielectric properties of the target, and $Y\colon\Omega\rightarrow\mathbb R_+$, the speckle noise, obeying a unitary mean Gamma law.
The distribution of the return, $Z=XY$, is completely specified by the distributions $X$ and $Y$ obey.
The univariate multiplicative model began as a single distribution, namely the Rayleigh law, was extended by~\citeasnoun{yueh89} to accomodate the $K$ law and later improved further by~\citeasnoun{frery96} to the $G$ distribution, that generalizes all the previous probability distributions.

The density function that describes the behavior of the speckle noise is
\begin{equation}
f_{X}(x)=\frac{L^{L}}{\Gamma(L)}{x}^{L-1}\exp\{-Lx\},L\geq1,x>0,
\label{gamma}
\end{equation}
where $L$ is the number of looks, a parameter related to the visual quality of the image and that can be controlled to a certain extent during the generation of the data.
An effective filter will tend to increase the value of this parameter, that when is estimated is referred to as ``equivalent number of looks''.

The most successful models for the backscatter are particular cases of the Generalized Inverse Gaussian distribution (Frery \textit{et al.}, 1997),  being the main ones a constant ($c$) and the Gamma $\Gamma(\alpha,\lambda)$, Reciprocal of Gamma $\Gamma^{-1}(\alpha,\gamma)$ and Inverse Gaussian $IG(\omega,\sigma)$ laws~\citeaffixed{harmoniceusar2000}{for this last see}.
These models for the backcatter yield the following distributions for the return $Z$, respectively:
\begin{description}
\item[Gamma:] characterized by the density function
\begin{equation}
 f(z)=\frac{{(L/c)}^{L}}{\Gamma(L)}{z}^{L-1}\exp\{-Lz/c\},
\label{densgamacons}
\end{equation}
with $c,z>0$ and $L\geq1$, denoted $\Gamma(L,L/c)$.
\item[K:] whose density function is $$f(z)=2\frac{(\sqrt {\lambda L})^{\alpha +L}  }{ \Gamma(L)\Gamma(\alpha) }z^{(\alpha+L)/2}K_{\alpha-L}(2\sqrt{\lambda Lz}),
$$
denoted $\mathcal K(\alpha,\lambda,L)$, where $z,\alpha,\lambda>0$, $L\geq1$, and $K_{\upsilon}$ is the modified Bessel function of third kind and order $\upsilon$;
\item[G0:] with density function
\begin{equation}
f(z)=\frac{L^{L}\Gamma(L-\alpha)}{\gamma^{\alpha}\Gamma(L)\Gamma(-\alpha)}\frac{z^{L-1}}{(\gamma+Lz)^{L-\alpha}},
\label{densga0}
\end{equation}
where $-\alpha,\gamma,z>0$, $L\geq1$, denoted $\mathcal G^{0}(\alpha,\gamma,L)$.
\item[GH:] characterized by the density function
\begin{equation}
\begin{split}
f(z)=\,&	2\frac{L^{L}\sqrt{\omega\sigma/\pi}}{\Gamma(L)e^{-2\omega}}
z^{L-1} \left(\frac{\omega}{\sigma(\omega\sigma+Lz)}\right)^{\frac{1+2L}{4}}\\
&K_{L+1/2}\left(2\sqrt{
\frac{\omega}{\sigma}(\omega\sigma+Lz)}\right),
\end{split}
\label{densh}
\end{equation}
where $\omega,\sigma,z>0$, $L\geq1$, denoted $\mathcal G^H(\omega,\sigma,L)$.
\end{description}

The use of these models in SAR image understanding has led to excellent results, as can be seen in the works by Mejail \textit{et al.}, 2003; Quartulli and Datcu, 2004. 

New Bayesian filters for retrieving information on $X$ based on the observation of outcomes of $Z$ will be presented in the next section.

\section{SPECKLE FILTERS}\label{sec:filtros}

One of the most widely used filters for speckle reduction is recalled, namely Lee's filter; then, the general setup for MAP filter is provided and two new Bayesian filters are derived.

\subsection {Lee's filter}

This filter~\cite{Lee86} aims at combating either multiplicative or additive noise or a combination of both.
It uses the observed mean and variance in a window to estimate the (unobserved) true backscatter by
$$
\widehat{x}=\overline{x}+b(z-\overline{x}) 
$$ 
where $b$ is an estimator of the ratio of the variance of $X$ to the variance of $Z$.
If no reliable model for $X$ is available, its moments have to be estimated from the data.
The mean and the variance of the backscatter can be estimated in every window by $
\overline{x}=\overline{z}/\overline{y}=\bar{z}$ and by 
$$\widehat{\sigma^2_X} = \widehat{\sigma^2_Z} - 
\frac{\overline{z}^{2}\sigma^2_Y}{(\sigma^2_Y+1)},
$$ where $\sigma_Y$ is the speckle standard deviation, that can be easily computed by means of the equivalent number of looks.

\subsection {MAP filters}

MAP (\textit{maximum a posteriori)} filters are a Bayesian approach to the problem of estimating the properties of $X$ given the observation of $Z$ by means of maximizing the posterior distribution of $X$ given $Z$
$$
f_{X\mid Z=z}(x)=\frac{f_{Z\mid X=x}(z)f_{X}(x)}{f_{Z}(z)}.
$$
Since $f_{Z}(z)$ does not depend on the sought variable $x$, the MAP estimator is defined as
$$
\hat{x}=\arg\max [f_{Z\mid X=x}(z)f_{X}(x)]
$$
or, since all quantities are positive, as
\begin{equation}
\label{xmap1}
\hat{x}=\arg\max[\log(f_{Z\mid X=x}(z))+\log(f_{X}(x))].
\end{equation}

In this work two MAP filters will be derived, those assuming the  $\Gamma^{-1}(\alpha,\gamma)$ and $IG(\omega,\sigma)$ laws for the backscatter. 
These filters will be called MAP $\mathcal G^0$ and $\mathcal G^H$.
Solving equation~\eqref{xmap1} in these two situations leads to the following estimators:
\begin{equation}
\label{xmapg0}
\hat{x}=\frac{Lz+\gamma}{L+1-\alpha}
\end{equation}
and
\begin{equation}
\label{xmapgh}
\hat{x}=\frac{(L+3/2)- \sqrt{(-L-3/2)^{2}+4\omega/\sigma(Lz+\omega\sigma)}}{-2\omega/\sigma},
\end{equation}
respectively.
The equivalent number of looks $L$ is estimated beforehand for the whole image using homogeneous areas~\cite{MejailJacoboFreryBustos:IJRS}. 
The parameters $\gamma$, $\alpha$, $\omega$ and $\sigma$ are estimated locally using the data available in a small window around the position being filtered; among the possible estimation techniques, in this work we used estimators based on the moments of order $1/2$ and $1$.
For maximum likelihood estimation and its numerical issues, the reader is referred to the work by~\citeasnoun{FreryCribariSouza:JASP:04}; improved inference by resampling is treated by~\citeasnoun{CribariFrerySilva:CSDA}.

The performance of these three filters is assessed in the following sections.

\section {LEE'S PROTOCOL}\label{sec:protocolo}

\begin{figure*}[htb]
\begin{center}
\subfigure[Phantom\label{img:uncorrupted:phantom}]{\includegraphics[width=.24\linewidth]{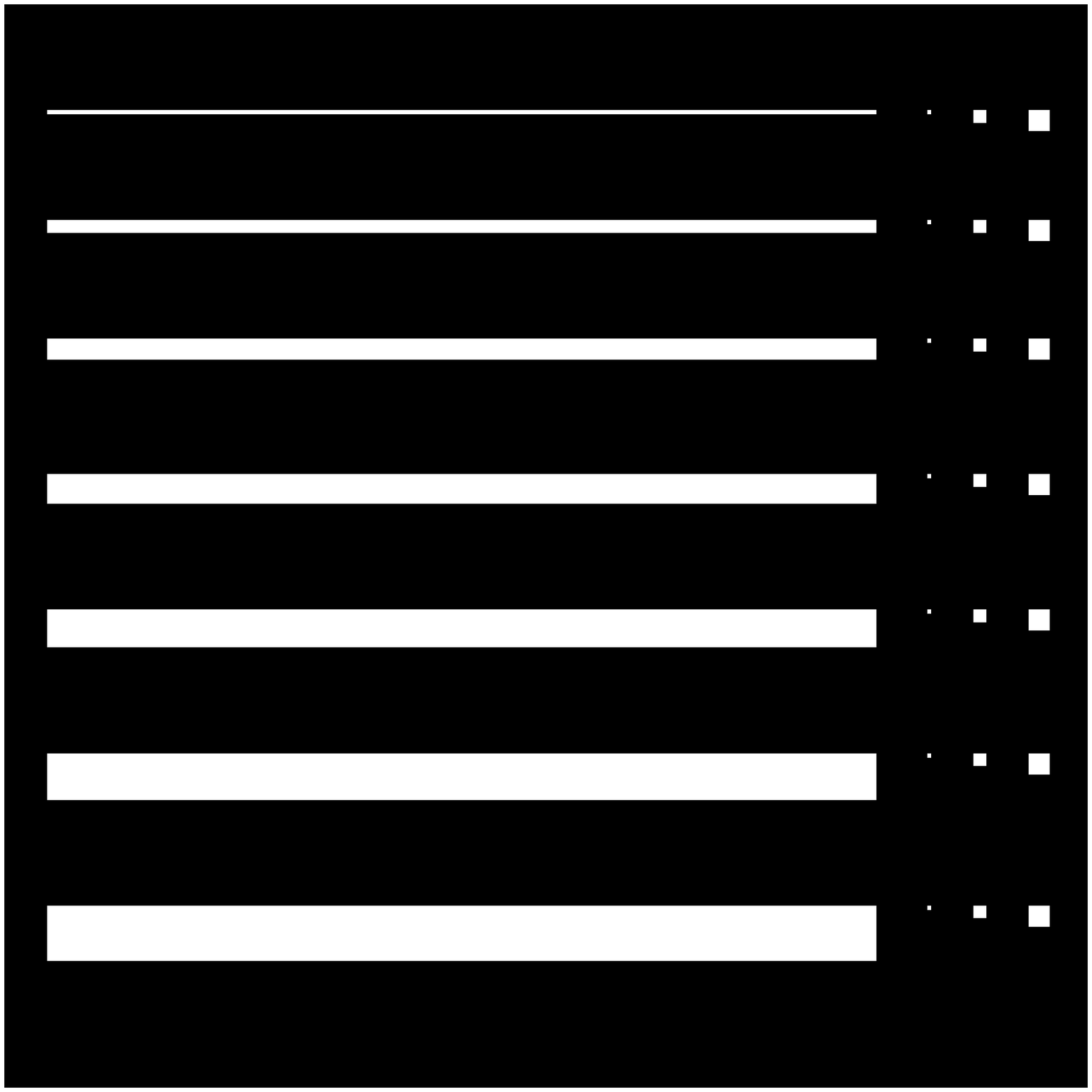}}
\subfigure[Corrupted phantom\label{img:speckle}]{\includegraphics[width=.24\linewidth]{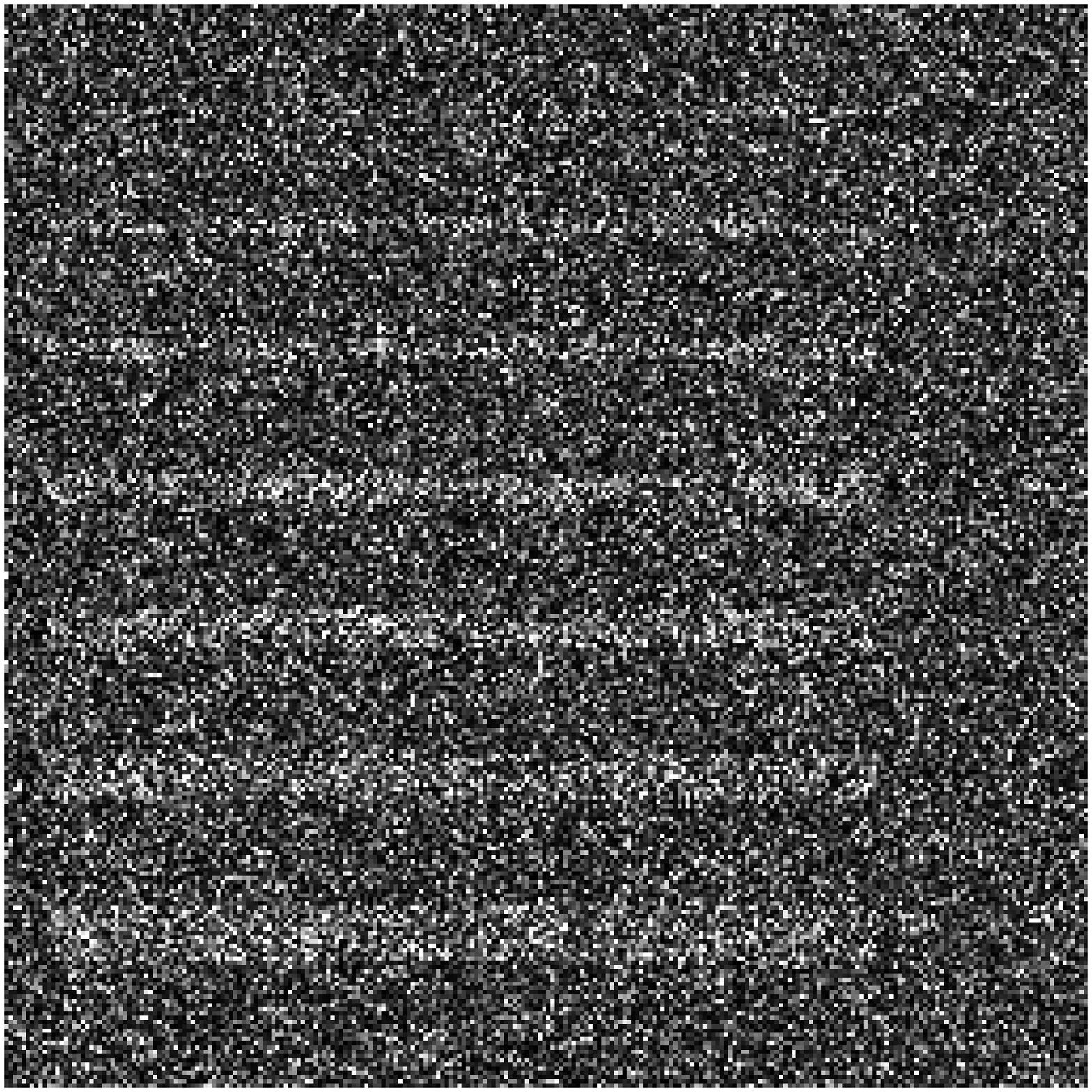}}
\subfigure[Lee filtered data\label{img:lee}]{\includegraphics[width=.24\linewidth]{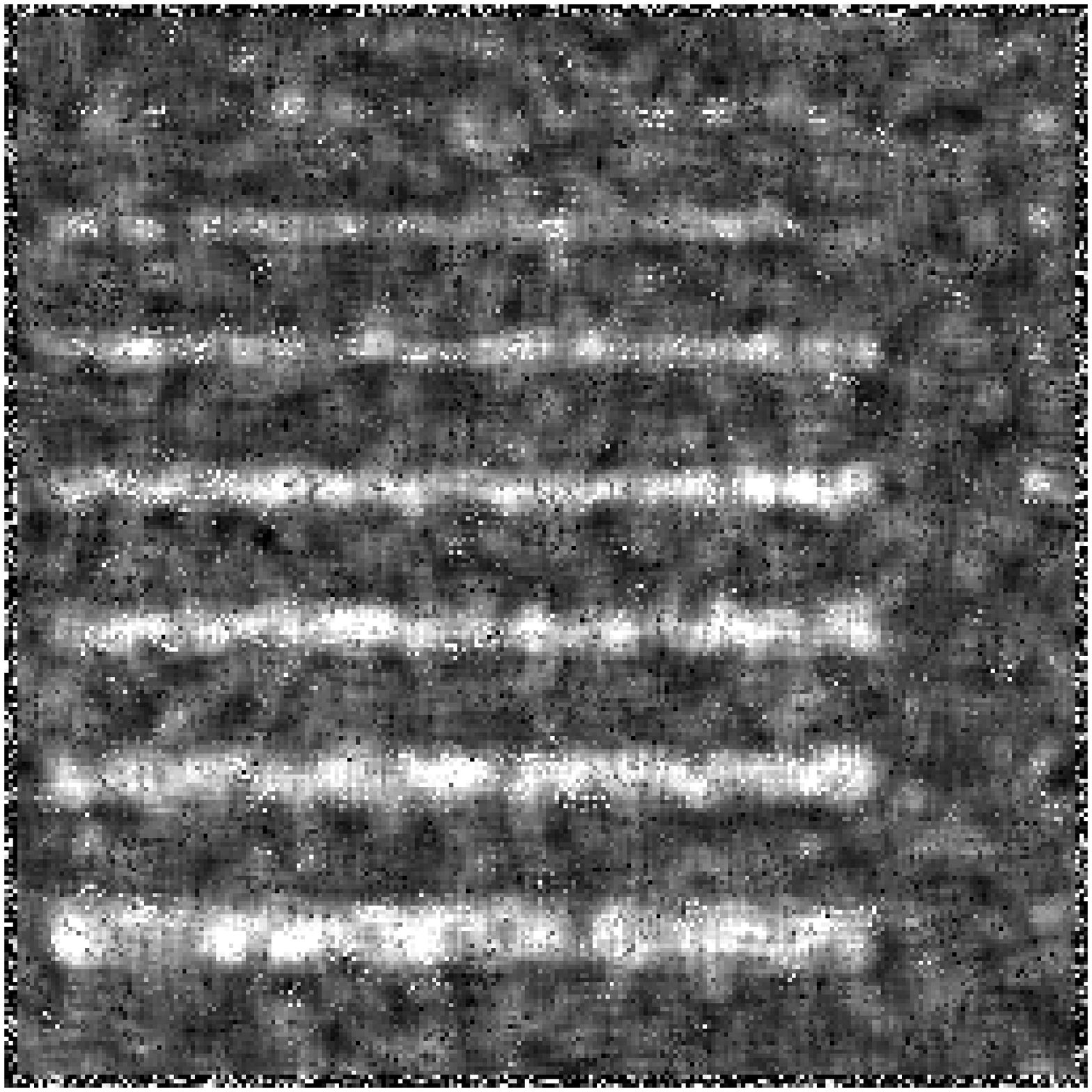}}
\subfigure[$\mathcal G^0$ filtered data\label{img:gi0}]{\includegraphics[width=.24\linewidth]{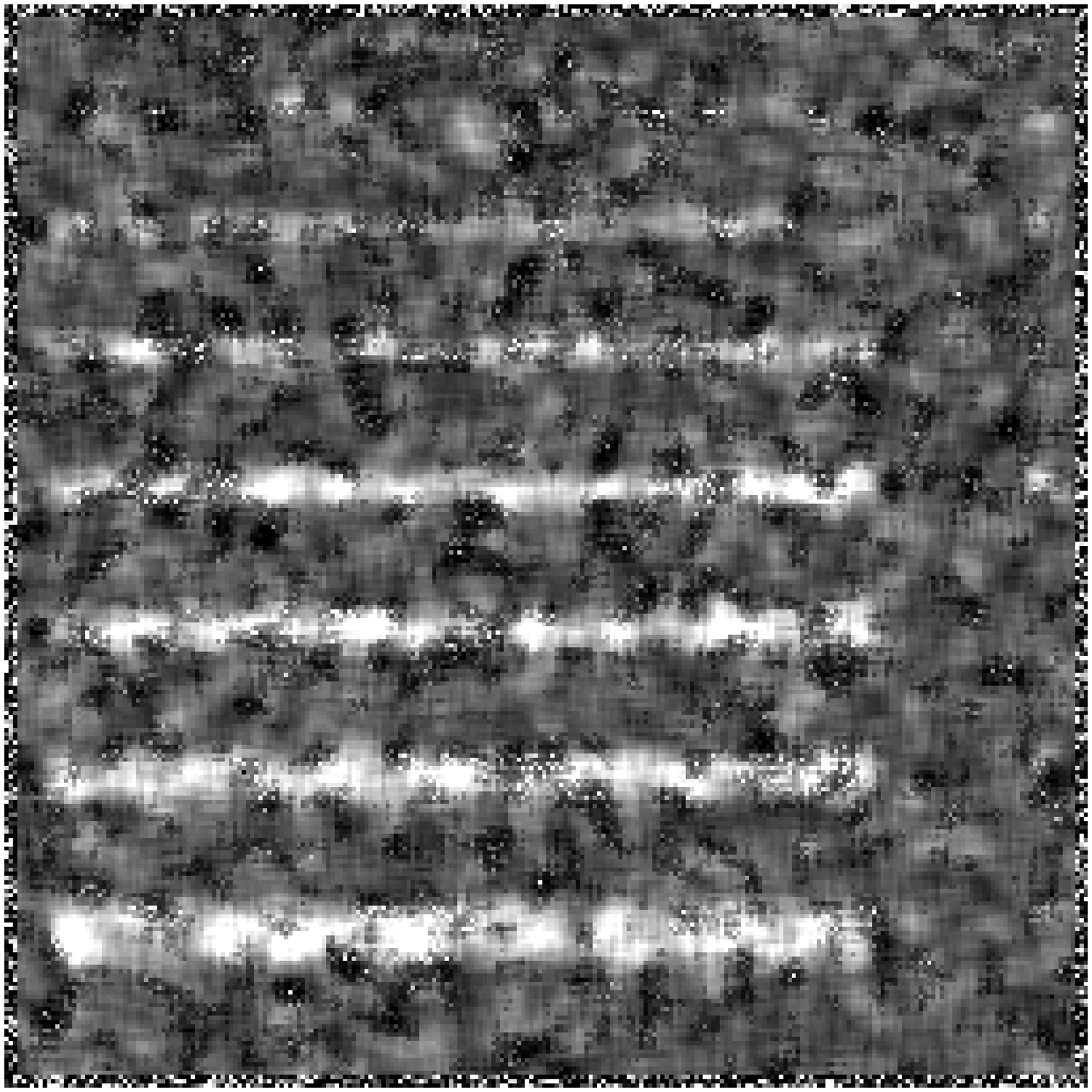}}
\end{center}
\caption{Lee's Protocol phantom, speckled data and filtered images}
\label{four:images}
\end{figure*}

Lee \textit{et.al}, 1994  proposed a protocol for the performance assessment of speckle reduction techniques.
This protocol consists of using a phantom image (see Figure~\ref{img:uncorrupted:phantom}) corrupted by speckle noise (see Figure~\ref{img:speckle}) and obtaining measures on the filtered versions; as an example, Figures~\ref{img:lee} and~\ref{img:gi0} present the results of two candidates for evaluation: Lee's and $\mathcal G^0$ filters.
Since the geometric properties of the uncorrupted phantom are known (points and strips of varying width from $1$ to $13$ pixels), it is possible to quantitatively assess the behavior of the techniques.

The equivalent number of looks, the line contrast and edge preserving are the used criteria to quantify the quality of speckle reduction techniques, determined by
\begin{description}
\item[Equivalent number of looks:] in intensity imagery and homogeneous areas, it can be estimated by $NEL=({\overline{z}}/{\widehat{\sigma_Z}})^2$, i.e., the square of the reciprocal of the coefficient of variation.

\item[Line contrast:] since the phantom has a line of one pixel width, the preservation of this line will be assessed by computing three means: those in the coordinates of the original line ($x_\ell$) and in those corresponding to two lines around it ($x_{\ell_1}$ and $x_{\ell_2}$).
The contrast is then defined as $2x_\ell - (x_{\ell_1}+x_{\ell_2})/2$, and compared with the contrast in the phantom image.

\item[Edge preserving:] it is measured by means of the edge gradient and variance. The former is computed as the absolute difference of the means of strips around edges, while the latter is as the former but using variances instead of means.
\end{description}

The best filter corresponds to the smallest value for all the measures, with exception of the Equivalent number of looks(in this case it occurs the opposit). 

It is noteworthy that the original proposal (see Lee \textit{et al.}, 1994) uses a single corrupted image to make the assessment.
This is a twofolded limitation, first related to the precision and secondly to the inherent variability of the measures, as we illustrate with an example.

A Monte Carlo version of Lee's protocol is proposed in next section for improved precision and significance, and the filters are then assessed with it.

\section{IMPROVED PROTOCOL FOR FILTER ASSEMENT}\label{sec:montecarlo}

The result of applying Lee's protocol to an image obtained by simulation is shown in Table~\ref{tab2}, where the ``best'' results are highlighted in boldface.
As can be seen, the four criteria lead to conflicting conclusions: 
\begin{itemize}
\item the best edge gradient is achieved by the $\mathcal G^0$ filter but, at the same time, it has the highest variance, which is detrimental;
\item the equivalent number of looks and the edge variance point at Lee's filter as the best, but it is the worst in line preservation;
\item if line preservation were to be used alone, the best filter is $\mathcal G^H$.
\end{itemize}

As previously seen, Lee's protocol preconizes the use of quantitative measures for the assessment of speckle-reduction techniques.
Their scale and relative importance are not of the same order, for example the difference in edge variance introduced by Lee's filter and $\mathcal G^H$ is $143.00-141.29=1.71$, while the equivalent number of looks produced by the two MAP filters differ by $19.04-17.24=1.80$, so it remains unclear how, if possible, to combine these results in a single conclusion.
Moreover, since no distributional assumption is made on these measures, one does not know anything about the significance of seemingly different results.

\begin{table}[htb]
\caption{Comparison of Speckle filters in a single image}\label{tab2}
\begin{center}
\begin{tabular}{crrrrr}\hline \hline
 Speckle   &\multicolumn{1}{c}{Edge} &\multicolumn{1}{c}{Edge}  & \multicolumn{1}{c}{$NEL$}  &\multicolumn{1}{c}{Line}\\ 
  \multicolumn{1}{c}{filter} &\multicolumn{1}{c}{Gradient} &\multicolumn{1}{c}{Variance} & &\multicolumn{1}{c}{Pres.}\\ \hline
  Lee&            $38.46$ & $\mathbf{141.29}$&     $\mathbf{25.17}$&   $79.88$&  \\
 $\mathcal G^0$&  $\mathbf{4.11}$ &  $164.15$&    $19.04$&   $11.30$&  \\
 $\mathcal G^H$&  $8.97$ &  $143.00$&    $17.24$&   $\mathbf{0.21}$&   \\
\hline\hline
\end{tabular}
\end{center}
\end{table}

An idea of precision of each measure, and not mere point estimation, becomes thus paramount in order to make fair filter assessments.
We propose the use of a Monte Carlo experiment as a means for obtaining this relevant information.

Such simulation experience will also help removing conflicting results as, for instance, the following situation observed for the same set of parameters and two simulated images $Z_1$ and $Z_2$.
Using the filtered versions of $Z_1$, one may conclude that Lee's filter is better than $\mathcal G^0$, since the equivalent number of looks they produce are, respectively, $28.32$ and $15.09$.
But if $Z_2$ filtered images are used, the same measures are $22.54$ and $29.88$, leading to the opposite conclusion.
Same conflicting results arise when other quality measures are applied, so having an idea of the variability of these factors is indispensably necessary.

Since all the observed data involved have a stochastic nature it is possible to devise Monte Carlo experiments.
Such \textit{in silico} experiment will allow the assessment of the performance of the MAP filters provided by equations~\eqref{xmapg0} and~\eqref{xmapgh}, avoiding the risk of basing one's decision on a single observation of a process.

It is necessary to design a Monte Carlo experiment with several situations in order to provide a global assessment of filter performance.
This experiment consists of simulating corrupted images as matrices of independent samples of the $\mathcal G^0$ distribution with  different parameters.
These parameters are based on previous experience with real data, and depict a few typical situations often encountered when analyzing SAR imagery.
These situations stem from a constant background (the original proposal, called ``Situation 0'') to extremely heterogeneous return (Situations 1 to 6).
Besides simulating different types of return, various contrasts between the dark background and the light foreground were also considered.
One hundred replications were performed for each of the situations depicted in Table~\ref{tab1}.

\begin{table}[htb]
\caption{Simulated situations with the $\mathcal G^{0}(\alpha,\gamma)$ distribution}\label{tab1}
\begin{center}
\begin{tabular}{crrr}\hline \hline
   Situation ID &$\alpha$ &\multicolumn{1}{c}{$\gamma$} &Background mean\\\hline
        1&                    $-2$&           $230$&       $230$          \\
        2&                    $-2$&            $50$&        $50$          \\
        3&                    $-4$&           $690$&       $230$          \\
        4&                    $-4$&           $150$&        $50$          \\
        5&                   $-10$&          $2070$&       $230$          \\
        6&                   $-10$&           $450$&        $50$          \\
\hline\hline
\end{tabular}
\end{center}
\end{table}

Every simulated image was subjected to Lee's and MAP $\mathcal G^0$ and $\mathcal G^H$ filters with windows of size $7\times 7$, and the comparison was perfomed by means of the criteria presented in Section~IV.


\section{RESULTS} \label{sec:resultados}

The results obtained are summarized by means of boxplots.
Each boxplot describes the results of one filter in a particular situation, using one hundred replications.

\begin{sidewaysfigure*}
\begin{center}
\subfigure[Equivalent Number of Looks\label{enl}]{\includegraphics[angle=270,width=.4\linewidth]{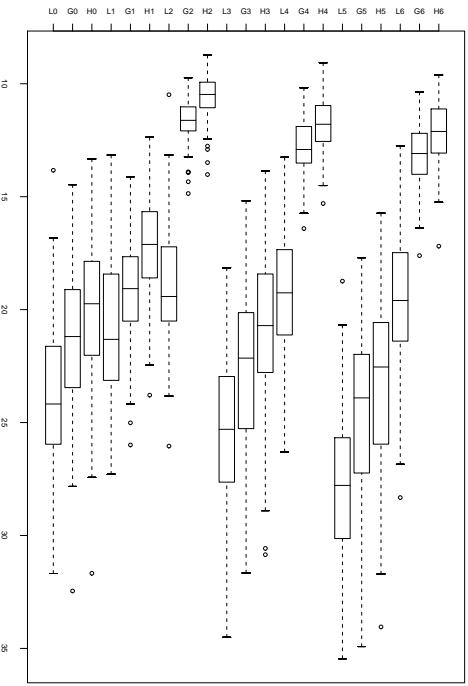}}
\subfigure[Line Preservation\label{line:preservation}]{\includegraphics[angle=270,width=.4\linewidth]{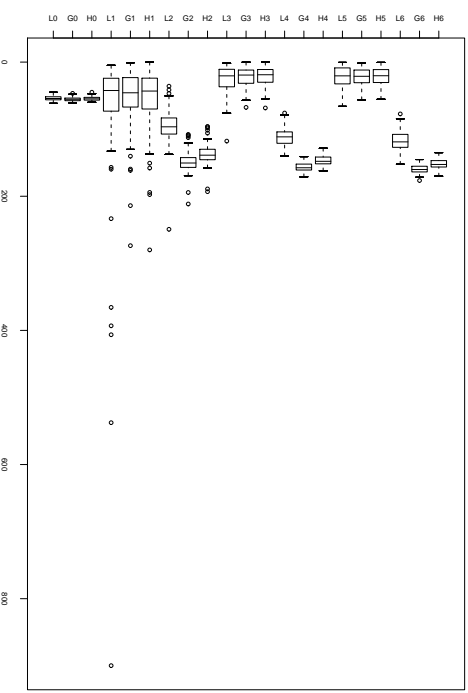}}
\subfigure[Edge Gradient\label{edge:gradient}]{\includegraphics[angle=270,width=.4\linewidth]{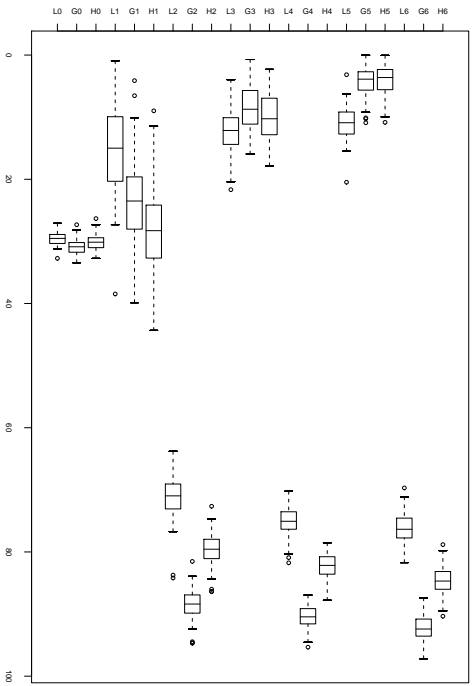}}
\subfigure[Edge Variance\label{edge:variance}]{\includegraphics[angle=270,width=.4\linewidth]{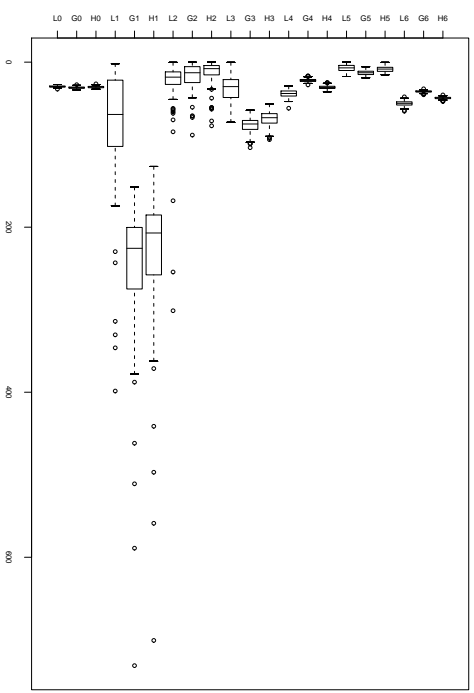}}
\end{center}
\caption{Boxplots of metrics applied to three filters in seven situations}
\label{fig:boxplots}
\end{sidewaysfigure*}

Figure~\ref{fig:boxplots} shows the boxplots of the four metrics corresponding to three filters in seven situations.
Vertical axes are coded by means of the filter (`L' for Lee, `G' for $\mathcal G^0$ and `H' for $\mathcal G^H$) and the situation (from $0$ to $6$).
All results exhibit a notorious variability with an exception: when both background and foreground truth are constant, i.e., when there is no backscatter variation.
Besides this variability, little can be said in order choose amongst the filters in the situations considered with the metrics proposed.

In the original situation (\#0), for instance, though there is little variability the values are almost the same for the four metrics.
In situation \#1 only edge variance is capable of making a discrimination, and Lee is the best filter (see Figure~\ref{edge:variance}).
For situation \#2, the Lee filter is the best one with respect to all criteria with the exception of edge variance; for this last criterion there is no difference among filters.
In situation \#3 only edge variance is capable of discrimination, pointing at Lee as the best filter.
Regarding situation \#4, Lee filter is the best with respect to line preservation, but the worst regarding edge variance.
In situation \#5 both MAP filters show improvement with respect to Lee in both edge gradient and edge variance.
Finally, when situation \#6 is considered, though there is no clear evidence, Lee is better than the other two in all criteria but one: edge variance.

One can conclude, then, that the original Lee's protocol is inadequate for filter comparison when realistic situations are considered, e.g, when either background or foreground or both vary.
More sensitive measures should be used to solve this limitation.
\citeasnoun{santanna95} proposes a means to mix different metrics in a single scalar allowing, thus, to assessing filters with a grade.
\citeasnoun{WangBovik:02} also propose a scalar metric for image quality assessment.

\section{CONCLUSIONS}\label{sec:conclusiones}

This paper presented two new MAP filters for speckle noise reduction and a Monte Carlo experiment that improves the original proposal.
When trying to assess their perfomance using this simulation setup, we noticed that this procedure is inadequate to deal with situations more realistic than constant background.

Previous works (see, for instance,~\citeasnoun{MedeirosMascaCosta:03};~\citeasnoun{santanna95} relate the superior behavior of MAP filters with respect to Lee's filter, but the high variability of the measures proposed by Lee's protocol hampers a straightforward comparison.

Other measures are showing a good potential for this assessment, such as the image index quality provided by~\citeasnoun{oliverquegan98} and the edge preserving index proposed by~\citeasnoun{Sattar:97}.

Monte Carlo experiments must be devised within the framework of statistical modelling in order to make sensible comparisons among filters.

\end{document}